# Leveraging Explainable AI to Analyze Researchers' Aspect-Based Sentiment about ChatGPT

Dr. Shilpa Lakhanpal[1], Dr. Ajay Gupta[2], Dr. Rajeev Agrawal[3]

[1]Department of Computer Science, California State University Fullerton, Fullerton, CA
[2]Department of Computer Science, Western Michigan University, Kalamazoo, MI
[3]Office of the Deputy Assistant Secretary of the Army (Procurement), ASA (ALT), Crystal City, VA

## Abstract

The groundbreaking invention of ChatGPT has triggered enormous discussion among users across all fields and domains. Among celebration around its various advantages, questions have been raised with regards to its correctness and ethics of its use. Efforts are already underway towards capturing user sentiments around it. But it begs the question as to how the research community is analyzing ChatGPT with regards to various aspects of its usage. It is this sentiment of the researchers that we analyze in our work. Since Aspect-Based Sentiment Analysis has usually only been applied on a few datasets, it gives limited success and that too only on short text data. We propose a methodology that uses Explainable AI to facilitate such analysis on research data. Our technique presents valuable insights into extending the state of the art of Aspect-Based Sentiment Analysis on newer datasets, where such analysis is not hampered by the length of the text data.

## 1   Introduction

ChatGPT [1] is a generative artificial intelligence (AI) chatbot which has revolutionized the landscape of Natural Language Processing (NLP). Up until the year 2022, NLP had made strides in various tasks such as text classification, question answering, summarization, sentiment analysis, named entity recognition, etc. But ChatGPT is the first of its kind to be released into the public domain, and perform a variety of these NLP tasks with effortless ease and be accessible to an ever-increasing user base. It is the fastest-growing consumer application in history, and it has set a record of reaching 100 million monthly active users in January 2023 within just 2 months of its release [2]. Comparatively, to reach a 100 million users, it took 9 months for TikTok, 2.5 years for Instagram, 5 years for Twitter, 7 years for World Wide Web and 75 years for the telephone. It is accessible as a question answering interface, where it provides intelligent, coherent and pertinent human-like answers to questions posed by the users, surpassing any other AI chatbot of its generation in popularity as depicted by Figure 1 [3].

---

Corresponding author: Dr. Shilpa Lakhanpal
Email: shlakhanpal@fullerton.edu

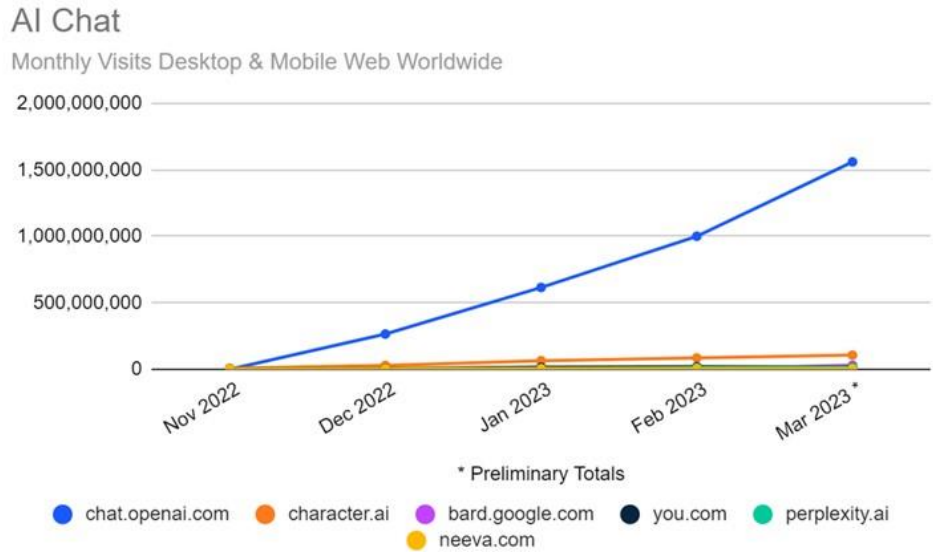

Figure 1

Ever since its release on November 30, 2022, ChatGPT has dominated public discourse with people taking to the social media and writing about their experience with ChatGPT. There are millions of tweets on Twitter. The public discourse ranges from sentiment of marvel to skepticism to apprehension as to the various levels of usefulness, applicability, and possibilities that ChatGPT is poised to bring in the near future [4, 5, 6]. The opinion paper [6] assimilates 43 contributions from experts in the fields of computer science, marketing, information systems, education, policy, hospitality and tourism, management, publishing, and nursing. These experts appreciate and recognize ChatGPT's potential to offer significant advantages to these fields benefitting their overall productivity. They also consider its limitations, possible disruptions it may cause to accepted practices, threats it may pose to privacy and security, and the consequences of biases, misuse, and misinformation arising from its use. Naturally, the euphoria surrounding ChatGPT has spurred the research community to investigate ChatGPT from various points of view including but not limited to how people across various sections of society view it and perceive its utility as, what ethical questions it raises, how the technology behind it can be improved etc. In fact, it is this sentiment about ChatGPT conveyed by the researchers in their papers that we investigate in our work. We look at how researchers across the world perceive ChatGPT. In doing so we investigate whether and how well some of the widely used sentiment analysis language models are able to capture the research community's exploration thus far. Particularly, since the focus of their research papers is ChatGPT, are the sentiment analysis language models able to capture their sentiment towards ChatGPT with aspect to various issues, fields and domains?

## 2    Sentiment Analysis in Research Articles

Sentiment analysis is an active field in NLP, where the goal is to identify the sentiment expressed in the text and classify it as a fixed polarity value such as positive, negative, or neutral. Sentiment analysis is used to extract sentiment from a wide arena of user bases such as social media and networking posts about products, state of current affairs, restaurants etc., or product sections in online retailer websites containing customer reviews, etc. Such analysis is important as for example, businesses can gain insight into customers' opinions about their products and accordingly adapt their product development and marketing strategies. While majority of efforts are focused on the classification and analysis of sentiment in customer reviews [7], and social media posts [8], lesser effort is dedicated to extracting and interpreting sentiment from research articles. With the success and now prolific use of transformers [9] and transformer-based models in various NLP tasks [10]; the task of sentiment analysis is also benefitted by these models. Even so, transformer-based models such as Bidirectional Encoder Representations from Transformers (BERT) [11] that are being used for sentiment analysis are still largely being used for analyzing social media content [12] or product reviews [13]. The use of research article data figures heavily in the field of biomedical research [14], where various tasks of NLP (not including sentiment analysis) are being performed by transformer-based models. Sentiment analysis lags behind in the field of biomedical research, because of lack of domain-specific sentiment vocabulary [15]. To the best of our knowledge, text data from Computer Science research articles has not been used for sentiment analysis, and in our work, we use such data.

## 3 Aspect-Based Sentiment Analysis (ABSA)

Largely sentiment analysis is performed either at the document-level or the sentence-level, where a single polarity value of either positive, negative or neutral is identified as the entire sentiment of the document or the sentence respectively. A single polarity value may not be an accurate sentiment representation of the entire document or the sentence as fine-grained sentiments may be need to be extracted towards several aspects in the document or sentence [16]. It is this Aspect-Based Sentiment Analysis (ABSA) that is gaining traction in the recent years. To understand what ABSA entails, let us look at the example laptop review in Figure 2. At the document-level, the entire text looks more positive than negative. At the sentence level, we can break the document into 5 sentences and classify each of them as positive, negative or neutral. At the aspect-level, we get information about the aspects such as *resolution*, *design*, *price* and *battery*. While the review is positive with respect to the former two aspects, it is negative towards the latter two. If we have 15 reviews about this laptop, 10 customers may find it expensive, while 5 may think it's okay. 14 customers may think the resolution is great, but 1 may not. Any product in general may have multiple aspects to be considered. This extends to restaurants, and even ChatGPT. Sometimes, the aspects are not explicitly mentioned, and can only be inferred. Such as the aspect *design* can be inferred by "*sleek and light*".

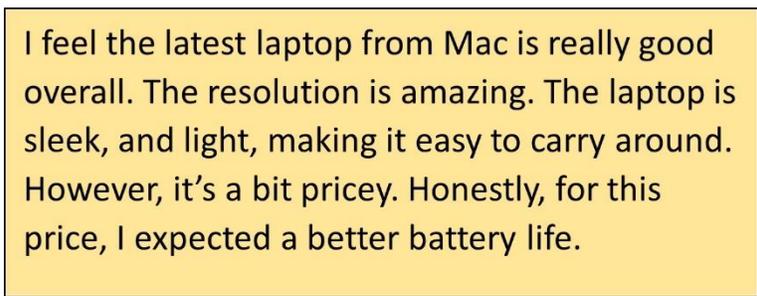

Figure 2

ABSA research involves sentiment elements such as aspect category, aspect term, opinion term and sentiment polarity [16]. Aspect categories need to be pre-defined such as food, service for the restaurant domain. Aspect term denotes the actual item mentioned or referred to, in the text such as *ice-cream* or *pizza* belonging to a food category. Opinion term is term used to convey feeling or sentiment about the aspect term. And sentiment polarity denotes whether the opinion is positive or negative, etc. For the restaurant domain, an example review and corresponding sentiment elements are presented in Figure 3.

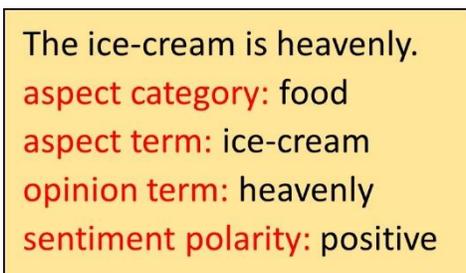

Figure 3

Many advancements have been done in aspect term extraction, aspect category detection, opinion term extraction, and aspect sentiment classification or some combinations thereof [16]. But only so many datasets have been extensively used. So far, the SemEval datasets [17, 18, 19] made public as part of shared work held during the International Workshop on Semantic Evaluation, held annually from 2014 to 2016, have been the most widely used for ABSA. Despite their popularity, most sentences in the dataset only include one or more aspects with the same sentiment polarity, effectively reducing the ABSA task to sentence-level sentiment analysis [20]. Also, these SemEval datasets contains reviews from only one domain, either restaurant or laptop. In [21], SentiHood dataset based on neighborhood domain is introduced, where work has been done to identify sentiments towards aspects of one or more entities of the same domain. With the use of transformer-based models, successful work [22] has been done to jointly detect all sentiment elements from a given opinionated sentence, for one or more aspects, even when aspects may have contrasting sentiments. The problem still remains that datasets used commonly have unique data structure and ABSA tasks successful on one dataset cannot be easily translated to another dataset.

A major step required in sentiment analysis is annotating the dataset for training the classifier. This step is expensive and often impossible in ABSA tasks as they require aspect-level annotations. To address this problem, transfer learning can be applied where we can leverage knowledge learned from one domain and apply on another. This involves taking a pre-trained language model such as BERT, which has been trained on a large dataset, and fine-tuning it for downstream tasks such as sentiment analysis. Fine-tuning requires labeled dataset which is specific to the task. Fine-tuned models of BERT exist for sentiment analysis such as:

- nlptown/bert-base-multilingual-uncased-sentiment [13], which we will refer to as the nlptown model

Fine-tuned models of BERT exist for ABSA such as:

- yangheng/deberta-v3-base-absa-v1.1 [23], which we will refer to as the yangheng model

These models can be further fine-tuned for ABSA tasks on target aspect-level labeled dataset. Which brings us back to the original problem of unavailability of labeled datasets. In the next section we describe our dataset.

# 4 Dataset

We collected data from arXiv [24] using the arXiv API [25]. The data comprises of metadata of research papers documenting research focused on ChatGPT, it's applications and implications. Specifically, we collected data from 868 papers submitted between December 8, 2022 and July 24, 2023 that contain the term chatgpt in either the title or the abstract or both.

The items from the metadata that we used for analysis are the titles and abstracts. For each paper, we added the title as a sentence before the abstract and refer to the resulting text again as the abstract for the corresponding paper.

## 4.1 Challenges in Analyzing Sentiment in this Dataset

For aspect terms to be extracted, much work is done in supervised learning, which requires labeled data. But research is still lacking in unsupervised learning [26]. Our dataset is unlabeled; hence it becomes difficult to extract aspects.

Most datasets [18, 19, 21] used in ABSA tasks are from the restaurant, laptop or neighborhood domain. Majority of the reviews in these contain only 1 sentence each, very few contain 2-3 sentences. All papers in our dataset are about ChatGPT, so we can qualify our domain as the ChatGPT domain. Each abstract in our dataset contains on an average 8 sentences, with the largest containing 19. Although models using transformers can capture some long-term dependencies, they may still struggle with long documents [27], where important context or sentiment clues are spread far apart. The model's ability to maintain relevant context over extended lines of text can impact its overall performance. In long text, sentiments towards aspects might change over time, leading to aspect-level sentiment shifts. Handling these dynamic shifts is still an ongoing challenge. Even when the sentiment is consistent, it may be expressed in a nuanced fashion and may need to be interpreted.

# 5 Using Explainable AI to Analyze Aspect-Based Sentiment

## 5.1 Example 1

In Figure 4, we present an abstract [28] from our dataset.

> Why Does ChatGPT Fall Short in Providing Truthful Answers? Recent advancements in Large Language Models, such as ChatGPT, have demonstrated significant potential to impact various aspects of human life. However, ChatGPT still faces challenges in aspects like truthfulness, e.g. providing accurate and reliable outputs. Therefore, in this paper, we seek to understand why ChatGPT falls short in providing truthful answers. For this purpose, we first analyze the failures of ChatGPT in complex open-domain question answering and identifies the abilities under the failures. Specifically, we categorize ChatGPT's failures into four types: comprehension, factualness, specificity, and inference. We further pinpoint three critical abilities associated with QA failures: knowledge memorization, knowledge recall, and knowledge reasoning. Additionally, we conduct experiments centered on these abilities and propose potential approaches to enhance truthfulness. The results indicate that furnishing the model with fine-grained external knowledge, hints for knowledge recall, and guidance for reasoning can empower the model to answer questions more truthfully.

Figure 4

This abstract indicates that this paper investigates why ChatGPT falls short in providing truthful answers, categorizing its failures into comprehension, factualness, specificity, and inference. It proposes enhancing truthfulness by furnishing the model with fine-grained external knowledge, knowledge recall hints, and reasoning guidance. We create a sentiment analysis pipeline for the nlptown model using the transformers library from HuggingFace [10]. We use it to classify the overall sentiment of the abstract. The result is depicted in Figure 5(a). nlptown model rates sentiment on a scale of 1 to 5 stars, with 1 indicating most negative, and 5 indicating most positive. The model gives this abstract a 1 star rating with 10.22% probability, a 2 star rating with 32.27% probability and a 3 star rating with 37.04% probability indicating a very negative-neutral range sentiment with a cumulative probability of 79.53%. Different models give different scores, and each model gives a different score for each label. We turn to Explainable AI (XAI) to understand how the models arrive at these scores in order to interpret the results. We use SHapley Additive exPlanations (SHAP) [29] to visualize the text to understand which words or phrases influenced the model's decision and whether they had a positive or negative effect leading to the resulting score.

```
[[{'label': '3 stars', 'score': 0.37044963240623474},
  {'label': '2 stars', 'score': 0.32270216941833496},
  {'label': '4 stars', 'score': 0.17089851200580597},
  {'label': '1 star', 'score': 0.10217782855033875},
  {'label': '5 stars', 'score': 0.033771809190511703}]]
```

Figure 5(a)

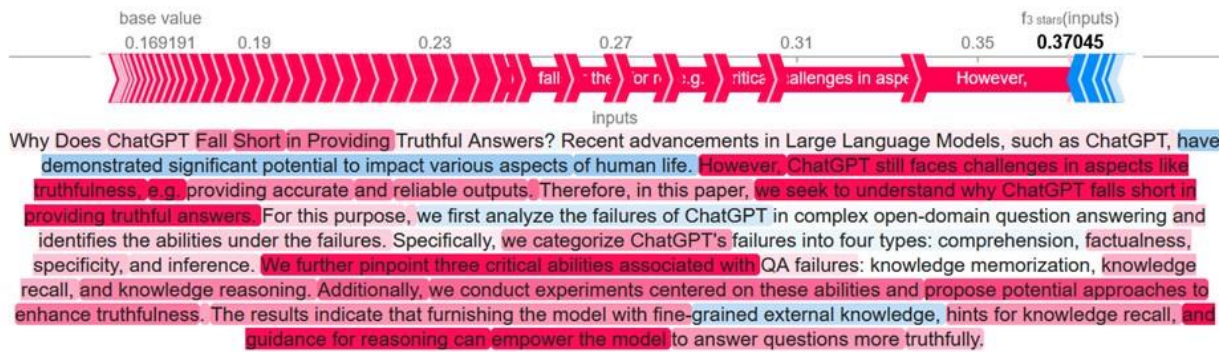

Figure 5(b)

Since the 3 stars (neutral) rating has the highest probability, in Figure 5(b) we present the SHAP text plot for the 3 stars rating. The words and phrases highlighted in red positively influence the model's prediction towards this rating, while the words and phrases in blue have an opposite effect. The darker colors indicate a stronger influence. In Figure 5(b), the text indicating that ChatGPT has demonstrated significant potential to impact human life is highlighted in blue, which means it contributes negatively towards the rating. The text indicating ChatGPT's failure to provide truthful answers or lack of truthfulness is highlighted in red in ALL instances. The text proposing approaches to enhance ChatGPT's truthfulness is also highlighted in red. The text in red pushes the model towards the 3 stars rating. **The resulting overall sentiment is *neutral (3)*.**

And ChatGPT agrees. We asked ChatGPT to label the overall sentiment of the abstract in Figure 4 on a scale of 1 to 5. Its response is shown in Figure 6.

The sentiment in the passage is neutral (3), as it objectively discusses ChatGPT's limitations in providing truthful answers while recognizing its potential impact. It identifies four types of failures and three critical abilities related to question answering shortcomings, followed by experiments and potential solutions for enhanced truthfulness.

Figure 6

Many sentiment analysis models only give rating as negative, neutral or positive. But since nlptown model rates at a more granular scale of 1 to 5, we can leverage it using Explainable AI for aspect term extraction tasks, which is otherwise impossible for datasets such as ours. And then we can perform ABSA using models such as the yangheng model. As a case in point, we apply the yangheng model to extract the sentiment of this abstract towards the aspect *truthfulness*. The results are depicted in Figure 7.

```
[[{'label': 'Negative', 'score': 0.6775800585746765},
  {'label': 'Positive', 'score': 0.24405549466609955},
  {'label': 'Neutral', 'score': 0.07836440950632095}]]
```

Figure 7

**It labels the text conveying a *negative* sentiment towards the aspect *truthfulness* with 67.76% probability**. Which is accurate since the text conveys that as it stands currently, ChatGPT lacks truthfulness. Hence, together both models can help understand the sentiment at a finer level.

## 5.2 Example 2

We present another abstract [30] from our dataset in Figure 8.

> A New Era of Artificial Intelligence in Education: A Multifaceted Revolution. The recent high performance of ChatGPT on several standardized academic test has thrust the topic of artificial intelligence (AI) into the mainstream conversation about the future of education. The objective of the present study is to investigate the effect of AI on education by examining its applications, advantages, and challenges. Our report focuses on the use of artificial intelligence in collaborative teacher-student learning, intelligent tutoring systems, automated assessment, and personalized learning. We also look into potential negative aspects, ethical issues, and possible future routes for AI implementation in education. Ultimately, we find that the only way forward is to accept and embrace the new technology, while implementing guardrails to prevent its abuse.

Figure 8

This abstract discusses the impact of AI in education and learning, focusing on its applications, benefits, and challenges. It highlights ChatGPT's success in academic tests and explores its role in collaborative teacher-student learning, intelligent tutoring systems, automated assessment, and personalized learning. The abstract acknowledges potential negative aspects and ethical concerns, but ultimately suggests accepting and embracing AI while implementing safeguards to prevent misuse.

We use the nlptown model to classify the overall sentiment of the abstract. The result is depicted in Figures 9(a).

```
[[{'label': '4 stars', 'score': 0.5352276563644409},
  {'label': '5 stars', 'score': 0.35541731119155884},
  {'label': '3 stars', 'score': 0.07598904520273209},
  {'label': '2 stars', 'score': 0.023732537403702736},
  {'label': '1 star', 'score': 0.009633398614823818}]]
```

Figure 9(a)

It indicates a positive – very positive range sentiment with a cumulative probability of 89.06%. In Figure 9(b) we present the SHAP text plot for the 4 stars (positive) rating.

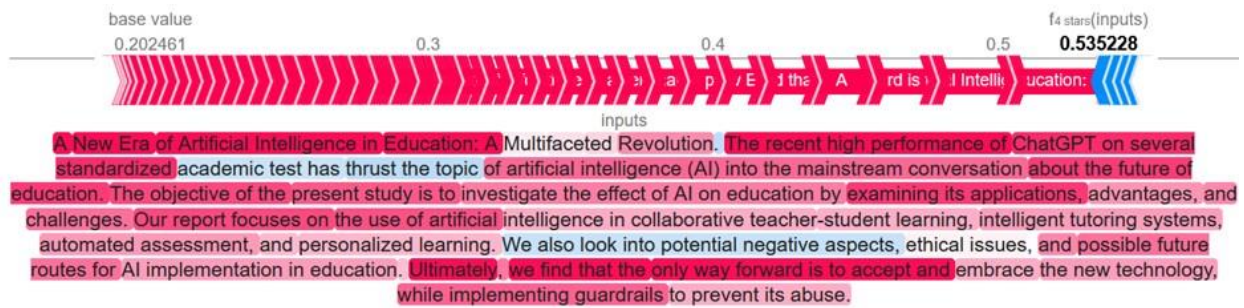

Figure 9(b)

The text has major portions recognizing the advantages and applications of ChatGPT in education and learning and these are highlighted in red, indicating that these portions contributed positively towards the 4 stars rating. The text about potential negative aspects is highlighted in blue. Hence the **overall rating is *positive (4)***.

ChatGPT agrees here as well and its response is shown in Figure 10.

The sentiment in the passage is positive (4), as it portrays the recent high performance of ChatGPT and AI in education as a promising development. It discusses various applications and advantages of AI in education, while also acknowledging the challenges and ethical concerns. The passage concludes with an optimistic tone, advocating for acceptance and embracing of the technology with responsible implementation to prevent misuse.

Figure 10

We find that the yangheng model **labels the sentiment of this abstract towards aspects such as *education* and *learning* as *positive***. The results are depicted respectively in Figure 11(a) and 11(b).

```
[[{'label': 'Positive', 'score': 0.5271556973457336},
  {'label': 'Neutral', 'score': 0.4476832151412964},
  {'label': 'Negative', 'score': 0.025161078199744225}]]
```

Figure 11(a)

```
[[{'label': 'Positive', 'score': 0.7254205346107483},
  {'label': 'Neutral', 'score': 0.2606089413166046},
  {'label': 'Negative', 'score': 0.013970516622066498}]]
```

Figure 11(b)

# 6 Conclusion

We use nlptown model to classify the overall sentiment about ChatGPT from each abstract in our dataset. In Figure 12, we show what percentage of abstracts convey the sentiment from 1 star to 5 stars. It can be seen that ChatGPT is well received by the researchers as 75.5% abstracts have a positive sentiment. The next highest sentiment is neutral at 11.8% followed by very positive at 6.1%. But as demonstrated above, these results should be interpreted with caution, as an overall sentiment may not correctly reflect the sentiment towards one or more aspects towards which ChatGPT may be applied. Particularly as we saw in Example 5.2, the overall sentiment and the sentiments towards the aspects under consideration was positive. But in Example 5.1, the overall sentiment was neutral but the sentiment towards the aspect under consideration was negative.

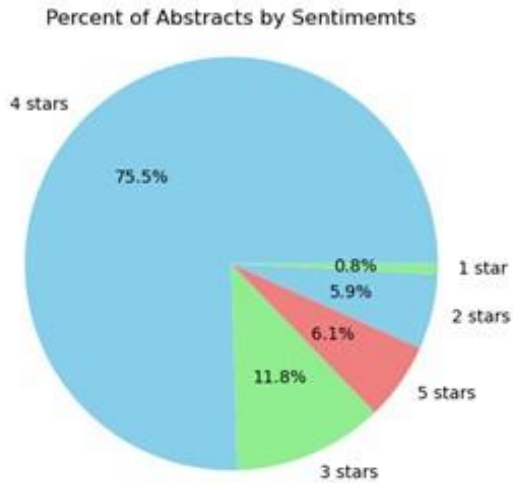

Figure 12

In Figure 13, we show what percentage of abstracts belong to which category in arxiv database. Arxiv database groups Computer Science papers under categories such as a cs.AI, cs.AR, etc. 50% of papers focused on ChatGPT belong to cs.CL (Computation and Language) category, 9% belong to cs.CY (Computers and Society), 6.6% belong to cs.SE (Software Engineering), 5.9% belong to cs.AI (Artificial Intelligence), etc. This indicates that researchers from various areas of computer science are analyzing the impact of ChatGPT impact towards various aspects.

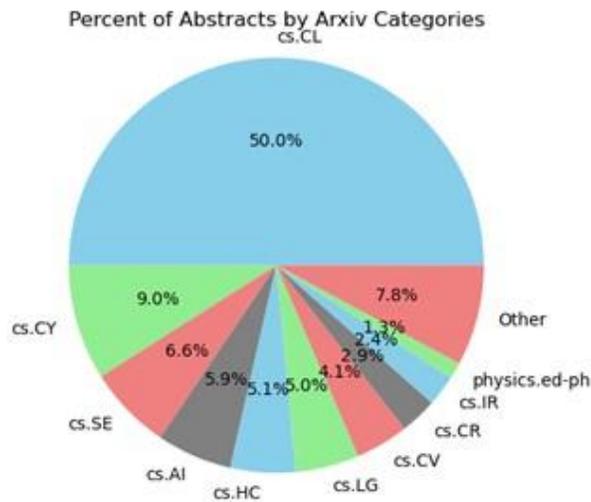

Figure 13

We have presented a technique for analyzing sentiments in Computer Science research articles. We found and demonstrated that the overall sentiment is not a correct representation of the sentiment of the entire text under analysis. Hence, we used Explainable AI to extract Aspect-Based Sentiment about ChatGPT towards various aspects. Our experiments present insights into how such aspects can be discovered from long texts using Explainable AI. As future work, we plan to investigate as to how we can detect the aspect terms as precisely as possible. In our work, we have extracted sentiment scores / labels using models such as the nlptown model and yangheng model. We have also validated the results by querying the ChatGPT model. As future work, we plan to apply and evaluate more models to improve the performance of Aspect-Based Sentiment Analysis.